\documentclass[10pt,conference]{IEEEtran}
\usepackage{cite}
\usepackage{amsmath,amssymb,amsfonts}
\usepackage{algorithmic}
\usepackage{graphicx}
\usepackage{textcomp}
\usepackage{xcolor}
\usepackage[hyphens]{url}
\usepackage{fancyhdr}

\usepackage[linesnumbered,ruled,vlined]{algorithm2e}
\usepackage{multirow}
\usepackage{pifont}
\usepackage{booktabs}
\usepackage{xcolor}
\usepackage{soul}
\usepackage{amssymb}
\usepackage{comment}
\usepackage[bookmarks=true,breaklinks=true,colorlinks,linkcolor=blue,citecolor=blue,urlcolor=black]{hyperref}

\newcommand{\squishlist}{
   \begin{list}{$\bullet$}
    { \setlength{\itemsep}{0pt}      \setlength{\parsep}{0pt}
      \setlength{\topsep}{3pt}       \setlength{\partopsep}{0pt}
      \setlength{\listparindent}{-2pt}
      \setlength{\itemindent}{-5pt}
      \setlength{\leftmargin}{1em} \setlength{\labelwidth}{0em}
      \setlength{\labelsep}{0.5em} } }

\newcommand{\squishend}{
    \end{list}  }

\newcommand{\todo}[1]{{\color{red}\sf\bfseries \hl{[TODO]} #1}}

\setuldepth{cat}
\newcommand{\note}[1]{{\color{magenta}$\square$}}

\newcommand{\hpcayear}{2026}

\title{Multi-objective Optimization in CPU Design Space Exploration: Attention is All You Need}

\author{\IEEEauthorblockN{\textbf{Runzhen Xue}\IEEEauthorrefmark{2}\IEEEauthorrefmark{3}, \textbf{Hao Wu}\IEEEauthorrefmark{4}, \textbf{Mingyu Yan}\IEEEauthorrefmark{2}\IEEEauthorrefmark{3}\IEEEauthorrefmark{1}, \textbf{Ziheng Xiao}\IEEEauthorrefmark{2}, \\ \textbf{Guangyu Sun}\IEEEauthorrefmark{5},\textbf{Xiaochun Ye}\IEEEauthorrefmark{2}\IEEEauthorrefmark{3}, 
\textbf{Dongrui Fan}\IEEEauthorrefmark{2}\IEEEauthorrefmark{3}}
\IEEEauthorblockA{
\IEEEauthorrefmark{2}State Key Lab of Processors, Institute of Computing Technology, Chinese Academy of Sciences\\ 
\IEEEauthorrefmark{3}School of Computer Science and Technology, University of Chinese Academy of Sciences\\ 
\IEEEauthorrefmark{4}University of Electronic Science and Technology of China\\
\IEEEauthorrefmark{5} School of Integrated Circuits, Peking University}
}

\fancypagestyle{camerareadyfirstpage}{%
  \fancyhead{}
  
  \fancyhead[C]{
    \ifdefined\aeopen
    \parbox[][12mm][t]{13.5cm}{\hpcayear{} IEEE International Symposium on High-Performance Computer Architecture (HPCA)}    
    \else
      \ifdefined\aereviewed
      \parbox[][12mm][t]{13.5cm}{\hpcayear{} IEEE International Symposium on High-Performance Computer Architecture (HPCA)}
      \else
      \ifdefined\aereproduced
      \parbox[][12mm][t]{13.5cm}{\hpcayear{} IEEE International Symposium on High-Performance Computer Architecture (HPCA)}
      \else
      \parbox[][0mm][t]{13.5cm}{\hpcayear{} IEEE International Symposium on High-Performance Computer Architecture (HPCA)}
    \fi 
    \fi 
    \fi 
    \ifdefined\aeopen 
      \includegraphics[width=12mm,height=12mm]{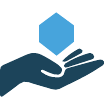}
    \fi 
    \ifdefined\aereviewed
      \includegraphics[width=12mm,height=12mm]{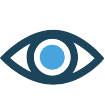}
    \fi 
    \ifdefined\aereproduced
      \includegraphics[width=12mm,height=12mm]{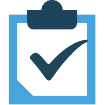}
    \fi
  }
  \fancyfoot[C]{}
}
\begin{document}
\maketitle

\ifdefined\hpcacameraready 
  \thispagestyle{camerareadyfirstpage}
  \pagestyle{empty}
\else
  \thispagestyle{plain}
  \pagestyle{plain}
\fi

\newcommand{\hpcaheight}{0mm}
\ifdefined\eaopen
\renewcommand{\hpcaheight}{12mm}
\fi


\renewcommand*{\thefootnote}{\fnsymbol{footnote}}
\footnotetext{$^*$Corresponding author is Mingyu Yan (yanmingyu@ict.ac.cn).}

\begin{abstract}


Design Space Exploration (DSE) is essential to modern CPU design, yet current frameworks struggle to scale and generalize in high-dimensional architectural spaces. 
As the dimensionality of design spaces continues to grow, existing DSE frameworks face
three fundamental challenges: 
(1) reduced accuracy and poor scalability of surrogate models in large design spaces;
(2) inefficient acquisition guided by hand-crafted heuristics or exhaustive search;
(3) limited interpretability, making it hard to pinpoint architectural bottlenecks.

In this work, we present \textbf{AttentionDSE}, the first end-to-end DSE framework that \emph{natively integrates} performance prediction and design guidance through an attention-based neural architecture. 
Unlike traditional DSE workflows that separate surrogate modeling from acquisition and rely heavily on hand-crafted heuristics, AttentionDSE establishes a unified, learning-driven optimization loop, in which attention weights serve a dual role: enabling accurate performance estimation and simultaneously exposing the performance bottleneck. This paradigm shift elevates attention from a passive representation mechanism to an active, interpretable driver of design decision-making.

Key innovations include:
(1) a \textbf{Perception-Driven Attention} mechanism that exploits architectural hierarchy and locality, scaling attention complexity from $\mathcal{O}(n^2)$ to $\mathcal{O}(n)$ via sliding windows;
(2) an \textbf{Attention-aware Bottleneck Analysis} that automatically surfaces critical parameters for targeted optimization, eliminating the need for domain-specific heuristics.

Evaluated on high-dimensional CPU design space using the SPEC CPU2017 benchmark suite, AttentionDSE achieves up to \textbf{3.9\% higher Pareto Hypervolume} and over \textbf{80\% reduction in exploration time} compared to state-of-the-art baselines.

\end{abstract}

\section{Introduction}

As modern CPUs continue to integrate more aggressive micro-architectural features and support increasingly diverse workloads, \textbf{design space exploration} (DSE) has become a crucial phase in the processor design cycle. DSE seeks to identify optimal configurations across a vast space of micro-architectural parameters under multiple conflicting objectives, such as performance, power, and area (PPA)~\cite{PPA1, PPA2, PPA3, PPA4}. To accelerate this process, prior work often formulates DSE as a multi-objective Bayesian Optimization (BO) task, leveraging a surrogate model to approximate PPA~\cite{PPA2, blackbox2, blackbox3, BOOMExplorer, MoDSE} and an acquisition function to guide design refinement~\cite{nonblackbox1, nonblackbox2, nonblackbox3, nonblackbox4, ActBoost, bottleneck_analysis1}.

However, with the escalating complexity of CPU micro-architecture, the dimensionality of the design space is growing rapidly---reaching hundreds of tunable parameters~\cite{PowerPC_cpu, AMD_cpu, superscalar_cpu}. This \textbf{high-dimensional design space} presents three critical challenges for existing DSE frameworks:

\squishlist
  \item \textbf{Inaccuracy and poor scalability of surrogate models:} Existing statistical models (e.g., GPR, ensemble trees) struggle to maintain accuracy and scalability in high-dimensional spaces~\cite{ActBoost, MoDSE, BO1, BO2, dimgpr, dimencurse}, suffering from sparsity of data and a long training time~\cite{random_forest, GPR}.
  \item \textbf{Ineffective acquisition functions:} Most acquisition strategies rely on heuristic rules~\cite{MoDSE, BOOMExplorer}, feature engineering~\cite{ArchExplorer}, or exhaustive search~\cite{MoDSE, BOOMExplorer, ActBoost}---which become impractical as the micro-architectural parameter count increases.
  \item \textbf{Lack of parameter-wise interpretability:} 
  Prior work offers limited visibility into the performance sensitivity or bottleneck contribution of each micro-architectural parameters~\cite{MoDSE, BOOMExplorer, ActBoost}, impeding informed design decisions.

\squishend

\noindent\textbf{Opportunity.} 
We observe that the above challenges are fundamentally rooted in the limited expressiveness and decoupled nature of existing DSE frameworks, where the surrogate model and acquisition function are treated as separate black boxes. 

Meanwhile, recent advances in attention-based deep learning—particularly transformers—offer a compelling alternative. First, attention mechanisms excel at modeling long-range dependencies across high-dimensional inputs, enabling them to accurately learn how diverse micro-architectural parameters jointly influence PPA. Second, the modular and parallelizable nature of attention layers ensures scalability as the number of micro-architectural parameters increases. 
Third, attention naturally learns to weight the influence of individual parameters when predicting target performance metrics, providing built-in interpretability that is essential for identifying architectural bottlenecks and guiding targeted design refinements.


These properties make attention-based models uniquely positioned to address the key challenges of DSE. By unifying surrogate prediction and acquisition within an attention-driven neural architecture, it becomes possible to achieve accurate, scalable, and explainable exploration—without relying on hand-crafted features or domain-specific rules. This convergence of modeling power, transparency, and scalability presents a timely opportunity to rethink DSE methodology for the era of high-dimensional CPU design.

\noindent\textbf{Our Proposal.}
To address the limitations of existing frameworks in high-dimensional CPU DSE, we propose \textbf{AttentionDSE}—an attention-driven framework that enables end-to-end, interpretable, and scalable DSE. AttentionDSE unifies performance prediction and exploration guidance within a single transformer-based neural architecture. Its predictor captures both the performance sensitivity and bottleneck contribution of each micro-architectural parameter, with attention weights serving a dual role: effectively supporting accurate PPA estimation and naturally exposing parameter-wise bottlenecks to directly guide the optimization process in a data-driven and interpretable manner.

To further enhance scalability and interpretability in high-dimensional settings, two key innovations are introduced:

\squishlist
\item \textbf{Perception-Driven Attention (PDA):}
A hardware-aware serialization and localized attention mechanism that captures architectural hierarchy and local connectivity, reducing attention complexity from $\mathcal{O}(n^2)$ to $\mathcal{O}(n)$ while greatly improving scalability without loss of modeling fidelity.

\item \textbf{Attention-aware Bottleneck Analysis (ABA):}
An exploration algorithm that interprets the attention weights to automatically identify performance-critical parameters, enabling fine-grained and explainable design updates without relying on expert knowledge or external heuristics.

\squishend

Together, these components enable AttentionDSE to form a closed-loop, data-driven DSE workflow that supports scalable, interpretable, and automated exploration in high-dimensional design spaces. This accelerates convergence and yields actionable insights into micro-architectural trade-offs—paving the way for more intelligent and insight-driven architecture design.

We evaluate AttentionDSE on the SPEC CPU 2017 benchmark suite using an OoO CPU with high-dimensional design space. Compared to state-of-the-art DSE frameworks (e.g., BOOMExplorer~\cite{BOOMExplorer}, MoDSE~\cite{MoDSE}, ArchExplorer~\cite{ArchExplorer}), AttentionDSE reduces exploration time by over 80\%, achieves up to 3.9\% improvement in Pareto Hypervolume (PHV), and maintains stable accuracy even in high-dimensional settings.

\noindent\textbf{Contributions.} 
This work makes the following contributions:

\squishlist
  \item We propose \textbf{AttentionDSE}, the first end-to-end attention-driven DSE framework that unifies performance prediction and exploration within a single interpretable architecture.

  \item We introduce \textbf{PDA}, a hardware-aware serialization and sliding attention mechanism that improves scalability in high-dimensional design spaces.

  \item We develop \textbf{ABA}, a novel method that uses attention weights for interpretable, automated performance bottleneck analysis and design refinement.
  
  \item We demonstrate significant improvements in accuracy, efficiency, and scalability over state-of-the-art DSE baselines on realistic CPU benchmarks.
\squishend

\section{Background}
In this section, we introduce the relevant concepts.

\subsection{Attention Mechanism}
The attention mechanism has become one of the most crucial concepts in deep learning. Inspired by human biological systems, which focus on distinctive and relevant details when processing large amounts of information, attention mechanisms allow models to prioritize important data instead of treating every input equally. This selective focus enhances the efficiency and effectiveness of handling complex data.

A typical attention layer in a neural network dynamically identifies the most relevant parts of the input. The attention score is defined as:
\begin{equation}
    \operatorname{Attention}(Q, K, V)=\operatorname{softmax}\left(\frac{Q K^{T}}{\sqrt{d_{k}}}\right) V .
\end{equation}
It operates using three key vectors: query (Q), key (K), and value (V). The attention mechanism calculates a score by comparing the query and key vectors, which are normalized via a softmax function to produce attention weights. These weights then compute a weighted sum of the value vectors, emphasizing the most critical information. 
This process is especially effective in self-attention, where each input element attends to all others, and in multi-head attention, which captures different aspects of the input through multiple attention heads. Attention layers are foundational in models like transformers~\cite{transformer}, enabling flexible and efficient learning of dependencies within sequences.

\subsection{Design Space Exploration in CPU Design}
DSE is a critical technique in the optimization and design process, primarily used to identify and analyze various design options within CPU design~\cite{PPA1, PPA2, PPA3, PPA4}. 
Its goal is to systematically explore all possible design alternatives to find the optimal or near-optimal solutions that meet specific design requirements and constraints~\cite{ArchExplorer, MoDSE, BOOMExplorer, ActBoost, wangduo1, wangduo2}. 

To expedite the process, prevailing DSE frameworks adhere to the fundamental principles of BO~\cite{MoDSE, BOOMExplorer, ActBoost, Archranker}.
BO leverages a \textbf{surrogate model} that serves as a predictive tool, approximating the underlying objective function~\cite{GPR, random_forest, ensemble_learning}. Additionally, an \textbf{acquisition function} drives the optimization process in DSE frameworks, efficiently navigating the DSE by identifying promising design points~\cite{PPA2}, which is a set of micro-architectural parameters, for evaluation~\cite{ArchExplorer}.
The above two facilitate effective exploration of the design space within a reasonable time~\cite{BO1, BO2}.


\begin{figure}[!htbp] 
	\centering
	\includegraphics[width=0.45\textwidth]{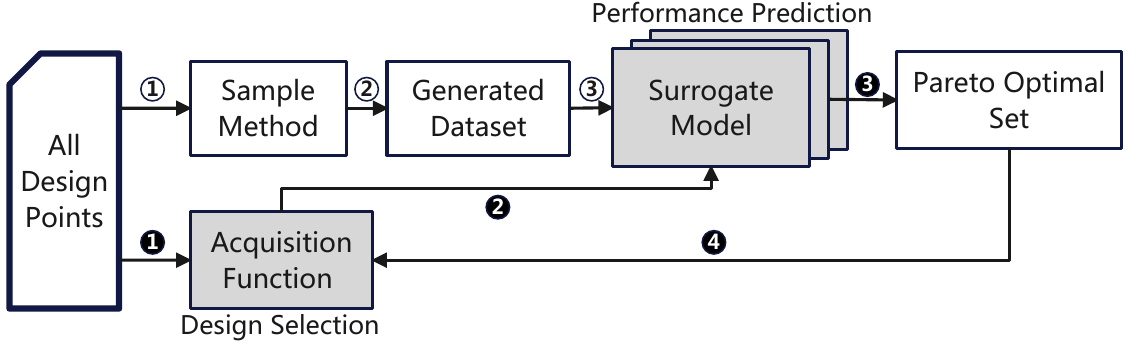}
	
	\caption{The prevailing DSE workflow. \ding{182}\~{}\ding{185} form the exploration loop.}
	\label{fig:DSE_framework_workflow}
\end{figure}

Fig.~\ref{fig:DSE_framework_workflow} illustrates the core workflow of the BO-based DSE framework. Prior to the exploration stage, the framework first trains the surrogate model using sampling methods, such as Plackett-Burman ranking\cite{PBrank, MoDSE}, to generate the training set  (\ding{172}, \ding{173}, \ding{174}). Once the surrogate models are trained, the exploration stage begins. The acquisition function selects a design point from the design space (\ding{182}) and sends it to the surrogate models to evaluate the PPA (\ding{183}). The framework then iteratively updates the Pareto-optimal set and repeats this process (\ding{184}, \ding{185}).

\subsection{Pareto Optimal Set}\label{sec:multi-objective_optimization}
A key concept in CPU DSE is the Pareto optimal set. A solution is considered as Pareto optimal if there is no other solution that improves some objective without worsening at least one other objective. 
To formalize the DSE as a multi-objective optimization task, the micro-architectural parameters are serialized as a feature vector $x$ referred to as a design point. The collection of all design points forms the design space $D$. Performance Metrics used to evaluate a design point are denoted as $y = f (x)$.
In an n-objective minimization problem, a design point $x$ is defined to be dominated by $x^{*}$ if
\begin{equation}
    \forall i \in[1, n], f_{i}\left(x^{*}\right) \leq f_{i}(x) ; \exists j \in[1, n], f_{j}\left(x^{*}\right)<f_{j}(x) .
\end{equation}
This dominance is denoted as the partial order $x^{*} \prec x$ and $y^{*} \prec y$. Conversely, if $x^{*}$ does not dominate $x$, it is denoted as $x^{*} \nprec x$ and $y^{*} \nprec y$. Among all design points, the set of design points that are not dominated by any other design points is called the Pareto optimal set 
$\Omega$, formulated as:
\begin{equation}
    \Omega=\left\{x \mid x^{*} \nprec x, \forall x^{*} \in D\right\} .
\end{equation}

\begin{figure}[!htbp] 
	\centering
	\includegraphics[width=0.4\textwidth]{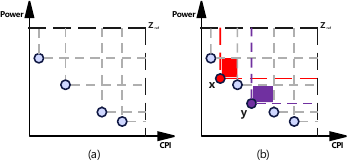}
	\caption{The improvement of PHV: (a) original PHV; (b) design points excelling in performance or power extend the PHV.}
	\label{fig:PHV_expend}
\end{figure}

The PHV is determined by the Pareto optimal set and a reference point, as shown in Fig.\ref{fig:PHV_expend}(a). It is represented by the region of blue points, enclosed by light gray lines, with the reference point labeled as $Z_{\text{ref}}$. A larger PHV indicates a better compromise achieved by the solution set across multiple optimization objectives. In Fig.\ref{fig:PHV_expend}(b), the new design points $x$ and $y$ demonstrate superior performance or power compared to the original design points, with both expanding the PHV, illustrated by the red and purple regions, respectively.

\section{Related Works}\label{sec:related_work}

Many efforts~\cite{Archranker, relative_work2, relative_work3, relative_work1, BOOMExplorer, MoDSE, ArchExplorer, relative_work4, ActBoost} accomplish the DSE tasks. 

For the early work, Mariani et al.~\cite{relative_work1} use a Kriging model~\cite{Kriging} with exponential correlation as a surrogate to predict the Pareto nondominated rank distribution. They then iteratively optimize using the Expected Improvement (EI) of non-dominated levels as the acquisition function.
CASH~\cite{relative_work2} compares multiple machine learning models and finds that BO achieves superior exploration results, by using a random forest as the surrogate model and EI as the acquisition function. 
Wang et al.~\cite{relative_work3} employ an adaptive component selection and smoothing operator as the surrogate model and the expected Hypervolume Improvement (HVI) of the Pareto frontier as the acquisition function. 

For the blackbox DSE framework, 
ActBoost~\cite{ActBoost} integrates statistical methods and machine learning to prune the design space and determine the training set. For the surrogate model, ActBoost uses two AdaBoost.RT~\cite{AdaBoost.RT} models with active learning. For the acquisition function, it calculates the prediction uncertainty of each unsampled design point and selects one with the highest uncertainty. By combining statistical sampling and the boosting technique, ActBoost enables efficient and accurate DSE tasks.
BOOMExplorer~\cite{BOOMExplorer} employs Gaussian Process Regression (GPR)~\cite{GPR} as the surrogate model and leverages EI~\cite{EI} as the acquisition function, searching for the next design point by analyzing data distribution characteristics. 
By embedding prior knowledge, BOOMExplorer achieves superior compared to analytical methods.
MoDSE~\cite{MoDSE} leverages ensemble learning for more accurate prediction. It combines AdaBoost~\cite{AdaBoost} and Gradient Boosted Regression Trees (GBRT)~\cite{GBRT} as the surrogate model and uses HVI with a full-space search for the acquisition function. MoDSE introduces a uniformity-aware selection algorithm for efficient multi-objective DSE and a Pareto-rank-based sample weight generation algorithm to train the surrogate model, enhancing the DSE outcomes.


For the DSE frameworks based on bottleneck analysis, ArchExplorer~\cite{ArchExplorer} adopts the dynamic event-dependence graph (DEG)~\cite{DEG} for critical path analysis and identifies bottlenecks in the current micro-architecture to guide the design of micro-architecture. It couples this approach with a software simulator (Gem5)~\cite{gem5} as the surrogate model. ArchExplorer focuses on refining the acquisition function to reduce the need for domain knowledge in mechanistic models.
Explainable-DSE~\cite{Explainable_DSE} is the SOTA framework for DSE tasks in deep neural network (DNN) accelerators. By leveraging the structure and characteristics of neural networks, Explainable-DSE constructs a bottleneck model to identify design bottlenecks. It enables targeted optimizations to mitigate issues such as high latency and power consumption, while also providing explanations for the design adjustments made and their impact on performance. This framework is primarily used for the joint DSE of DNN accelerators and DNN algorithms.



\section{Motivation}\label{sec:motivation}

In this section, we will explain the motivations behind leveraging the attention mechanism in the DSE framework.

\subsection{Challenges in Current DSE Studies}



As modern CPUs continue to integrate increasingly aggressive micro-architectural features and support a wider range of workloads, their design complexity grows substantially. This complexity is reflected in the expanding number of tunable micro-architectural parameters—spanning pipeline width, core frequency, and so on. As a result, the design space has become significantly larger and more intricate, giving rise to the challenge of high-dimensional DSE.

Existing DSE frameworks, however, struggle to scale in such settings and face three fundamental challenges in high-dimensional exploration tasks.

\textbf{First, the effectiveness of DSE frameworks is limited by the low accuracy and poor scalability of surrogate models in high-dimensional CPU design spaces.}
Most existing approaches~\cite{ActBoost, MoDSE, BOOMExplorer, Archranker} use statistical models—such as linear regression or Gaussian processes—to predict PPA. While efficient on small search spaces, these models quickly break down as the number of design parameters grows. For instance, when the design space exceeds 75 parameters, BOOMExplorer struggles to maintain prediction accuracy.

This is mainly due to the ``curse of dimensionality'': as the number of parameters increases, the design space becomes exponentially sparser, requiring significantly more samples to train reliable models~\cite{dimencurse}. Unfortunately, increasing training size leads to excessive overhead. For example, in Gaussian Process-based models, training requires matrix inversion with $O(n^3)$ complexity and inference scales as $O(n^2)$, where $n$ is the number of training samples. This makes it impractical to scale traditional surrogate modeling to realistic, high-dimensional CPU design scenarios.




\textbf{Second, the performance of DSE frameworks is hindered by the inefficiency of acquisition functions in high-dimensional design spaces.}
Acquisition functions are critical for guiding exploration toward optimal micro-architectural configurations. However, in large parameter spaces, existing strategies become inefficient and poorly scalable. Many methods depend on domain-specific heuristics~\cite{ActBoost, MoDSE, BOOMExplorer} or handcrafted features~\cite{ArchExplorer} to identify performance bottlenecks. These approaches are often biased, time-consuming, and difficult to generalize across architectures and workloads.

For example, ArchExplorer~\cite{ArchExplorer} analyzes only the first 100,000 instructions and takes 15 days to optimize an architecture with just 21 parameters. Some frameworks resort to large-scale or even exhaustive search~\cite{MoDSE, ActBoost}, which becomes computationally infeasible as dimensionality increases. \textsc{MoDSE}~\cite{MoDSE}, for instance, operates over a 10-parameter space with 36,864 design points, but scaling to 30 parameters would require an astronomical $1.9 \times 10^{29}$ years of search time.

These limitations highlight the need for acquisition strategies that can operate efficiently in high-dimensional settings—without relying on heuristics or exhaustive enumeration—while still identifying high-quality designs.



\textbf{Third, existing DSE frameworks lack interpretability, making it difficult to reason about trade-offs and locate architectural bottlenecks.}
Conventional frameworks offer little insight into how individual micro-architectural parameters affect PPA. This makes it challenging to understand why a given design performs well or poorly, thereby hampering informed, targeted design refinements.

This problem stems from two core limitations. First, most surrogate models rely on either mathematical analysis~\cite{ActBoost, Archranker} or black-box predictors~\cite{BOOMExplorer, MoDSE}, which fail to expose parameter-wise contributions to performance.
Second, the standard two-stage DSE workflow treats performance prediction and design exploration as separate steps~\cite{BOOMExplorer, MoDSE, ActBoost, ArchExplorer}, preventing feedback from the predictor from guiding future exploration. As a result, it is difficult for the acquisition function to adaptively prioritize parameters that are more performance-sensitive.

Without a unified, feedback-driven approach, existing methods are not only harder to interpret, but also less effective at navigating high-dimensional design spaces-especially when timely, bottleneck-aware decisions are required.

\subsection{Why Attention for DSE?}

The attention mechanism provides a powerful and well-suited solution to above challenges for several reasons:

\squishlist

\item \textbf{Context-aware Interaction Modeling:} Attention dynamically computes the relative importance of each parameter with respect to others in a given design configuration~\cite{attention4glpbal_relation}. This enables the model to capture complex, conditional dependencies—such as how the benefit of a wide fetch stage may depend on the reorder buffer size or instruction window depth—without manual feature engineering.

\item \textbf{Scalability and Structural Flexibility:} With recent advances such as sparse attention, attention-based models can scale to hundreds of parameters while maintaining modeling fidelity. Moreover, attention can be adapted to reflect architectural hierarchies and locality, enabling better alignment with the structure of CPU design spaces.

\item \textbf{Unified Modeling for Prediction and Guidance:}
Attention can bridge surrogate modeling and exploration by using a single predictor to both estimate PPA and inform the search direction. This eliminates  the need for hand-crafted heuristics or separate analysis stages and enabling a more efficient, closed-loop DSE process.

\item \textbf{Interpretability through Attention Weights:} Unlike black-box predictors, attention mechanisms expose internal weights that reveal which parameters contribute most to performance metrics. This provides a natural interface for bottleneck analysis and informed refinement, supporting more targeted and efficient exploration.

\squishend

These properties position attention not merely as a prediction tool, but as a unified interface between surrogate modeling and acquisition strategy. When properly integrated, attention mechanisms can enable an end-to-end, self-driven DSE framework that is both interpretable and scalable—paving the way for a new class of intelligent architecture design tools.

\section{Design of AttentionDSE}\label{}


In this section, we introduce AttentionDSE, an attention-based framework designed for high-dimensional DSE.

\subsection{Overview of AttentionDSE}\label{sec:overview_of_AttentionDSE}

\begin{figure}[!t] 
	\centering
	\includegraphics[width=0.45\textwidth]{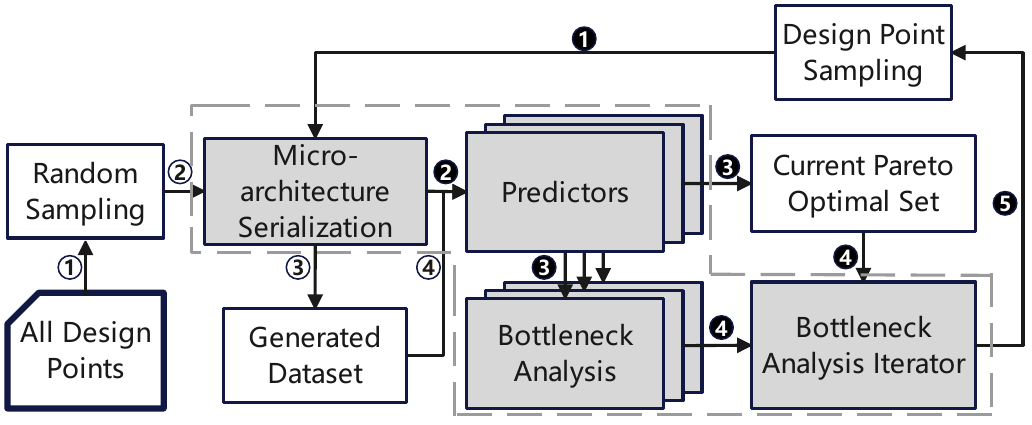}
	
	\caption{Overview of the AttentionDSE. }
	\label{fig:overview_of_AttentionDSE}
    \vspace{-5pt}
\end{figure}

Fig.~\ref{fig:overview_of_AttentionDSE} provides an overview of AttentionDSE. 
It comprises three primary components: \textbf{(1) Predictors}, which provide rapid and accurate predictions of performance metrics (e.g., IPC) for given micro-architectures across benchmarks; 
\textbf{(2) Micro-architecture Serialization}, designed to convert various design parameters into input vectors for the predictor while ensuring the positional certainty of parameters; it also reduces the computational burden associated with predictions, thereby expediting the performance evaluation process; \textbf{(3) Bottleneck Analysis and Iterator}, which identify the micro-architectural parameters that introduce performance bottlenecks within micro-architecture modules, facilitating the exploration of promising micro-architectural candidates by selectively modifying these parameters. The process of the AttentionDSE consists of a predictor training stage (\ding{172}, \ding{173}, \ding{174}, \ding{175}) and an architecture exploration stage (\ding{182}, \ding{183}, \ding{184}, \ding{185}, \ding{186}).

\begin{figure*}[!htbp] 
	\centering
	\includegraphics[width=0.9\textwidth]{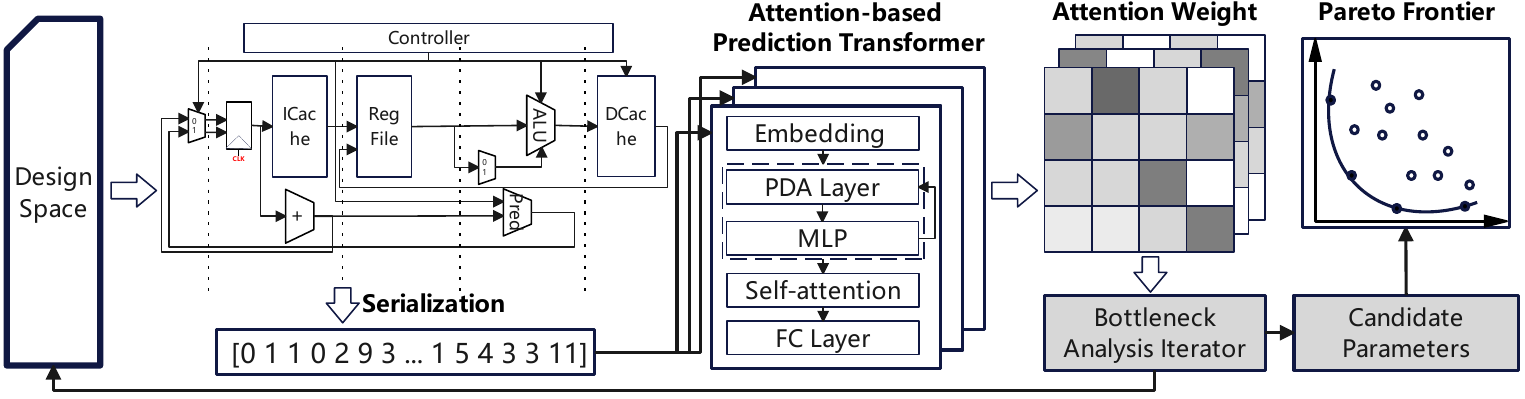}
	
	\caption{The workflow of AttentionDSE framework in the exploration stage.}
	\label{fig:unidse_exploration_workflow}
    \vspace{-5pt}
\end{figure*}

The training stage is a one-time process, and the sequence of operations is represented by hollow numbers. This stage focuses on training predictors for performance metrics, ensuring that the models accurately capture the relationships between micro-architectural parameters and performance metrics.
Unlike other approaches that rely on specialized sampling methods, AttentionDSE adopts a random sampling approach and then converts the design point to the sequence input, generating the training set (\ding{172}, \ding{173}, \ding{174}).
Next, the deep learning model, integrated with the PDA, functioning as the predictor, is trained to predict performance metrics (\ding{175}).
Benefiting from probabilistic attribute correlation in attention calculation, it more easily captures the intricate relationship between parameters, informing hints for the exploration stage.

Once the predictors are trained, AttentionDSE is ready for the exploration stage, which is iterative and aims for efficient multi-objective optimization. 
The exploration stage begins with a random sampling of design points (\ding{182}), followed by the exploration of the Pareto optimal set of these points through the trained predictors (\ding{183}). Each design point in the Pareto optimal set is analyzed to determine the contribution of each parameter (\ding{184}). The Bottleneck Analysis Iterator then selects the next design point based on the ABA algorithm (\ding{185}, \ding{186}). This iterative process continues until the preset performance metrics are met, ensuring efficient and thorough exploration of the design space.
Fig.~\ref{fig:unidse_exploration_workflow} presents a detailed workflow of the exploration stage. All these techniques will be discussed in detail in the following sections.

\subsection{Attention-based Prediction Model}\label{sec:single-objective_predictor}

A fast and accurate performance predictor is essential to replacing traditional simulation-based methods, forming the foundation of an efficient DSE framework. Such predictors enable rapid estimation of performance metrics, significantly accelerating design iterations and reducing overall exploration time. As discussed in Section~\ref{sec:motivation}, attention-based neural networks are particularly well-suited for this task: they not only excel at modeling complex, context-dependent relationships in high-dimensional design spaces, but also naturally support parallel computation, making them highly compatible with GPU-based acceleration. Thus, we introduce an attention-based model tailored for single-objective performance prediction. By combining strong modeling capacity with scalable inference, the proposed surrogate supports fast and accurate exploration across large micro-architectural design spaces.

\begin{algorithm}[!t]
    \SetAlgoLined
    \fontsize{8pt}{8pt}\selectfont
    \caption{\small \textbf{The Training Procedure of Predictors}}
    \label{alg:training_predictor}
    \KwIn{$D$: Design space; $n$: The size of the training set; $k$: Embedding length; $Depth$: The depth of the model; $Epoch$: The total epochs for training.}
    \KwOut{$M$: The trained predictor.}
    
    $T \leftarrow RandomSample(D, n)$; \\
    Instantiate the prediction token $T$ with embedding length $k$;\\
    Initialize the weights of the Predictor;\\
    Simulate $T$ to obtain $S_{ground\_truth}$ of IPC, Power, and Area;\\
    \For{$epoch \leftarrow 0$ \KwTo $Epoch$}{
        $S_{embed} \leftarrow Embedding(T)$;\\
        $S_{embed} \leftarrow Add(S_{embed}, PE)$;\\
        \For{$depth \leftarrow 0$ \KwTo $Depth$}{
            $S_{attn} \leftarrow SelfAttention(S_{embed})$;\\
            $S_{mlp} \leftarrow MLP(S_{attn})$;\\
        }
        $P_{t}, Attn\_{score} \leftarrow SelfAttention(S_{mlp})$;\\
        Extract analysis token $A_t \leftarrow Reduce\_{sum}(Attn\_{score})$;\\
        $S_{pred} \leftarrow FullyConnectedLayer(P_{t})$;\\
        $BackwardPropagation(S_{pred}, S_{ground\_truth})$;\\
    }
    return Predictor;\\
\end{algorithm}

Algorithm~\ref{alg:training_predictor} outlines the training procedure of our attention-based predictor. The process begins by randomly sampling $n$ design points from the design space $D$ and simulating each to obtain ground truth metrics (IPC, Power, and Area).
This procedure comprises the following steps.


\squishlist
\item Embedding: Each discrete, heterogeneous microarchitectural parameter is mapped to a vector representation, enabling the model to reason over diverse parameter types in a unified latent space.

\item Position Embedding (PE): Positional encodings are added to retain the identity and ordering of each parameter, allowing the attention mechanism to associate learned importance scores with specific microarchitectural features.
\squishend


\squishlist
\item Prediction Token ($P_t$): Prediction token aggregates global information via self-attention, and is processed by an MLP to produce final PPA predictions.

\item Analysis Token ($A_t$): Derived from attention scores from the final self-attention layer, $A_t$ captures the relative importance of each parameter, enabling bottleneck analysis and interpretability.

\squishend

By embedding structural context and aligning parameter positions, the model effectively captures complex interdependencies among architectural parameters—addressing the key challenge of scalability and insight in high-dimensional DSE.

\subsection{Perception-driven Attention  Mechanism}\label{sec:PDA}

Although the prediction time is significantly reduced compared to simulation, further reductions are still necessary, especially when handling the large-scale design spaces of CPUs.
Fortunately, from a micro-architectural design perspective, not all micro-architectural parameters are strongly correlated with one another, presenting an opportunity to reduce the computational burden of the attention mechanism significantly. For instance, the fetch buffer is closely related to the fetch width and fetch queue but has only a weak relationship with the writeback width. Therefore, in this section, we introduce the PDA mechanism as an alternative to the vanilla self-attention mechanism to alleviate the computational burden.
The PDA mechanism includes two aspects: the micro-architecture serialization method and the sliding window attention mechanism.

In previous work~\cite{GPR, MoDSE, ensemble_learning, BOOMExplorer}, the order of parameters is not a concern. Because GPR~\cite{GPR, BOOMExplorer} relies on the distance and similarity between input data rather than their arrangement order in statistical regression methods. Similarly, in ensemble learning~\cite{ensemble_learning, MoDSE}, the training process of multiple weak learners introduces randomness, making the order of parameters irrelevant.
To address this challenge in the attention-based predictor, we introduce a serialization approach termed Perception-driven Serialization (PDS), which preserves architectural information and establishes a foundation for the sliding window attention technique. This method leverages a graph-based framework to map relationships among parameters within each pipeline stage, rooted in a key insight: CPU instruction and data flows are processed sequentially across pipeline stages, with parameters within the same stage interacting far more frequently than those across different stages.

\begin{figure}[!htbp] 
	\centering
	\includegraphics[width=0.45\textwidth]{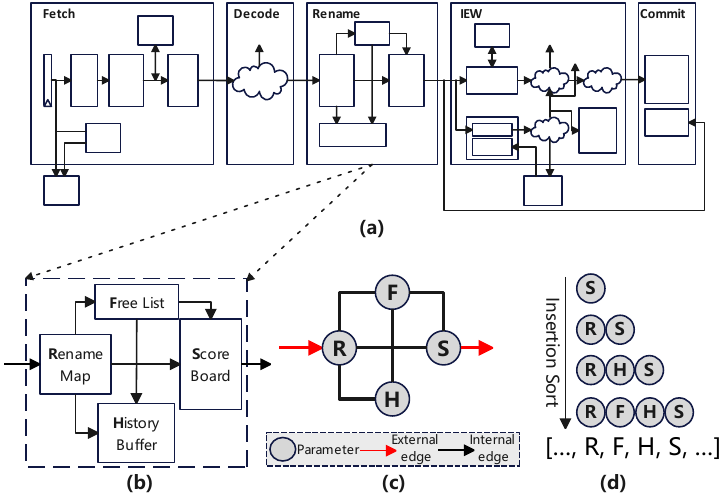}
	
	\caption{The Perception-driven Serialization approach. (a) A Sample OoO CPU core; (b) Rename Stage; (c) Stage abstraction; (d) The steps of Perception-driven Serialization approach.}
	\label{fig:micro-architecture_serialization_technique}
    \vspace{-5pt}
\end{figure}

An example is illustrated in  Fig.~\ref{fig:micro-architecture_serialization_technique}(a). We chose the Rename stage, shown in Fig.~\ref{fig:micro-architecture_serialization_technique}(b), as an example, where there are four parameters to be serialized, including the free list (F), the rename map (R), the scoreboard (S), and the history buffer (H). 
The PDS first transfers the micro-architecture into a perceptual graph, which is an undirected graph based on the dataflow among these components, as depicted in Fig.~\ref{fig:micro-architecture_serialization_technique}(c).
In a perceptual graph, vertices represent the parameters within each pipeline stage, while edges denote the datapaths between these components. The edges are categorized into two types: red edges, which connect to other pipeline stages and are labeled as external edges, and black edges, which connect to components within the current pipeline stage and are labeled as internal edges.
To quantify the perception of each component, we leverage the perception degree $D$ of each parameter in the perceptual graph. $D$ is calculated as follows:
\begin{equation}
    D(\text{parameter}) = \sum_{e \in E_{\text{parameter}}^{\text{internal}}} e - \sum_{e \in E_{\text{parameter}}^{\text{external}}} e
\end{equation}
Here, the internal edges represent the intra-stage perception field, while the external edges represent the inter-stage perception field.
After calculating the perception degrees, the sequence of the micro-architecture is determined using the insertion sort algorithm. Parameters with the larger perception degree are inserted into the middle of the sequence as shown in Fig.~\ref{fig:micro-architecture_serialization_technique}(d).
After all the pipeline stages have been serialized, the final vector of the micro-architecture is combined according to the order of the pipeline stages. 

\begin{figure}[!htbp] 
	\centering
	\includegraphics[width=0.45\textwidth]{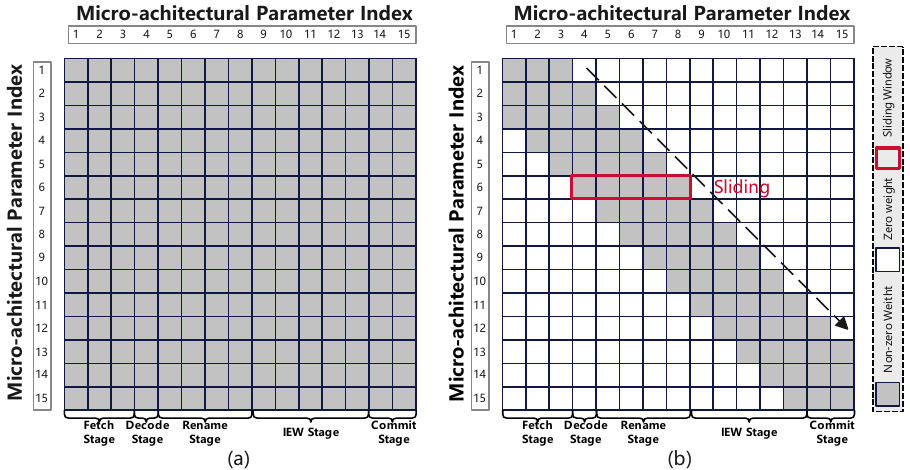}
	
	\caption{The attention weight calculation. (a) Vanilla self-attention mechanism; (b) PDA mechanism.}
	\label{fig:sliding_window_attention}
    \vspace{-5pt}
\end{figure}

In the clustered sequence, the correlation between adjacent parameters is significantly higher than that between non-adjacent parameters. At this point, the sliding window only needs to cover a local area to capture most of the key dependencies, without the need to calculate global attention.
Fig.~\ref{fig:sliding_window_attention} shows an example of attention weight calculation compared with vanilla self-attention and PDA. For vanilla self-attention, every parameter must calculate the attention weight with every other parameter, as shown in Fig.~\ref{fig:sliding_window_attention}(a).
To optimize this process, we set the window size to the maximum perception degree among all parameters. In this example, the window size is set to five, as illustrated in Fig.~\ref{fig:sliding_window_attention}(b). 

In general, with the help of the PDA, each parameter calculates the attention weight only with its relative parameters. In this way, for a parameter vector of length $n$, the computational complexity is reduced from $O(n^2)$ to $O(n)$. Thus, the PDA approach alleviates the limitations posed by the number of parameters, enhances scalability, and reduces the training and inference time of the prediction model.


\subsection{Attention-aware Bottleneck Analysis}\label{sec:ABA}


Bottleneck analysis is a crucial step in CPU design, enabling architects to identify the key microarchitectural factors that significantly influence performance. 
To support this capability, we propose the ABA algorithm (Algorithm~\ref{alg:HEA}), which captures intricate relationships between microarchitectural parameters and performance metrics. By identifying the parameters that contribute most to performance variation, ABA effectively guides DSE by prioritizing high-impact parameters, thereby improving both exploration efficiency and design quality.

To enable such analysis in an interpretable and quantifiable manner, we extract a dedicated analysis token $A_t$ from the previously introduced prediction model. This token is designed to assess the importance of each input parameter with respect to performance outcomes. A central requirement for ensuring the reliability of this mechanism is maintaining a strict one-to-one correspondence between attention weights and the input parameters. To ensure this, the model preserves token position consistency across all components: an embedding layer, PDA layer, MLP, self-attention, and fully connected (FC) layer. 
Specifically, the embedding layer applies independent projections to each parameter token, the PDA layer adds positional encoding while preserving order, and both the MLP and self-attention layers operate token-wise to retain the direct mapping between tokens and their associated attention weights.

This rigorous preservation of token alignment is vital: it enables each attention weight to be directly and explicitly mapped back to a specific design parameter. Leveraging this property, the ABA algorithm interprets the attention-weight heatmap generated by the model to identify performance-critical parameters. In this heatmap, each row indicates how much a particular parameter is influenced by others, while each column—indexed by the analysis token $A_t$—reflects the degree to which a parameter affects all other parameters. The column-wise sum of attention weights thus serves as a proxy for a parameter’s global impact on performance prediction.

Based on this insight, ABA detects bottleneck parameters by analyzing $A_t$: a low value for IPC (indicating high sensitivity) or a high value for Power or Area (indicating dominant cost contribution) signifies that the corresponding parameter is a bottleneck for that specific objective. ABA then updates the design configuration accordingly, refining the Pareto frontier and facilitating iterative improvements in both design quality and exploration efficiency.

\begin{algorithm}[!htbp]
    \SetAlgoLined
    \fontsize{8pt}{8pt}\selectfont
    \caption{\small \textbf{Attention-aware Bottleneck Analysis}}
    \label{alg:HEA}
    \KwIn{$D$: The design space; $X$, $X_{n}$: Initial design point set and its size; $I_{max}$: The max iteration of exploration;}
    \KwOut{$\Omega$: The Pareto optimal set}
    
    $X \leftarrow RandomSample(D, X_{n})$; \\
    $\Omega \leftarrow GetParetoOptimal(Predictor(X))$;\\
    \For{$i$ $\leftarrow$ 0 to $I_{max}$}{
        Set search queue $Q \leftarrow \varnothing$;\\
        \For{each design point x in $\Omega$}{
        $A_t \leftarrow Predictor(x)$;\\
        $x' \leftarrow BottleneckAnalysis(x, A_t)$;\\
        $PPA \leftarrow Predictor(x')$;\\
            \If{$x'~has~expended~the~\Omega$}{
                $Q.push(x')$;\\
            }
        }
        $\Omega \leftarrow GetParetoOptimal(Predictor(Q))$;\\
    }
    return $\Omega$;\\
\end{algorithm}


\begin{figure}[!htbp] 
	\centering
	\includegraphics[width=0.42\textwidth]{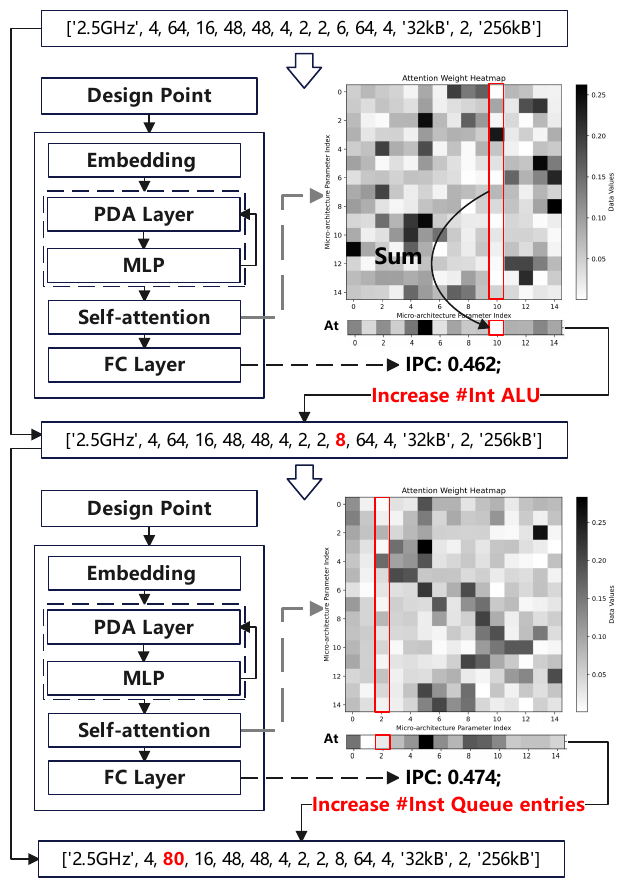}

	\caption{The example of ABA algorithm updating the parameters. }
	\label{fig:HEA_example}
    \vspace{-5pt}
\end{figure}

To clarify the execution process of the ABA algorithm, Fig.~\ref{fig:HEA_example} illustrates a toy example that demonstrates how the ABA algorithm updates micro-architectural parameters step by step to enhance performance (IPC). Initially, AttentionDSE randomly samples a design point and sends the parameters to the predictor. Based on the attention weight heatmap from the self-attention layer, the ABA algorithm sums all the columns and selects the parameter with the minimal sum as the current performance bottleneck, which in this example is the integer ALUs. The ABA algorithm then increases the number of ALUs from six to eight, improving performance.
In the subsequent iteration, the ABA algorithm identifies the instruction queue size as the current bottleneck using the same process. The ABA algorithm increases the queue size from 64 to 80. By strategically adjusting the instruction queue size, the pipeline is provided with a sufficient flow of instructions, further optimizing performance.

\section{Experimental Setup}

In this section, we introduce the experimental setup used to assess the performance of AttentionDSE.

\textbf{Simulation.}
We utilize GEM5~\cite{gem5} as the timing-accurate simulator and McPAT~\cite{mcpat} as the power and area modeling tool. 
McPAT is configured with technology parameters aligned with industry-standard VLSI design rules, enabling it to estimate power (dynamic and leakage) and area based on microarchitectural specifications, which is consistent with prior DSE works~\cite{MoDSE, BOOMExplorer}.
We extend GEM5 using Python to generate different CPU cores with varying configurations.
To bolster the credibility of our experiments, we employ SPEC CPU2017~\cite{SPEC2017} as the evaluation benchmark. For identifying resource utilization status, we utilize Simpoints~\cite{simpoints} for each workload evaluation. Each workload is divided into at most 30 clusters, with each cluster containing ten million instructions.

\textbf{Baselines.}
Our baselines include the ActBoost~\cite{ActBoost}, the BOOMExplorer~\cite{BOOMExplorer}, and the MoDSE~\cite{MoDSE}. 
Additionally, we implement ArchExplorer~\cite{ArchExplorer} for comparison with our ABA algorithm. The baselines represent recent SOTA work. All experiments are conducted on the Out-of-Order (OoO) CPU implemented by Gem5.
The design space of the OoO processor is listed in Table~\ref{tb:design_space}. The values in the third column are formatted as ``start number: end number: stride". A two-level cache hierarchy with an 8192MB DRAM main memory configuration is employed. The size of the design space exceeds $6.89 \times 10^{35}$, making exhaustive search impractical.

\begin{table}[!htbp]
\centering
\fontsize{8pt}{8pt}\selectfont
\caption{Micro-architecture design space specification.}
\renewcommand{\arraystretch}{1.2}
\label{tb:design_space}
\begin{tabular}{|c|cc|}
\hline
\textbf{Parameters} & \multicolumn{1}{c|}{\textbf{Description}}                                                                                    & \textbf{\begin{tabular}[c]{@{}c@{}}Candidate\\ Value\end{tabular}} \\ \hline
Core Frequency      & \multicolumn{1}{c|}{\begin{tabular}[c]{@{}c@{}}the frequency of\\ CPU core in GHz\end{tabular}}                              & \begin{tabular}[c]{@{}c@{}}1/1.5/2/\\ 2.5/3\end{tabular}           \\ \hline
Pipeline Width      & \multicolumn{1}{c|}{\begin{tabular}[c]{@{}c@{}}fetch/decode/rename/\\ dispatch/issue/writeback/\\ commit width\end{tabular}} & 1:12:1                                                             \\ \hline
Fetch Buffer        & \multicolumn{1}{c|}{\begin{tabular}[c]{@{}c@{}}fetch buffer size\\ in bytes\end{tabular}}                                    & 16/32/64                                                           \\ \hline
Fetch Queue         & \multicolumn{1}{c|}{\begin{tabular}[c]{@{}c@{}}fetch queue size\\ in $\mu$-ops\end{tabular}}                                 & 8:48:4                                                             \\ \hline
Branch Predictor      & \multicolumn{1}{c|}{predictor type}                                                                                          & \begin{tabular}[c]{@{}c@{}}BiModeBP/\\ TournamentBP\end{tabular}   \\ \hline
Choice Predictor    & \multicolumn{1}{c|}{choice predictor size}                                                                                   & 2048/4096/8192                                                     \\ \hline
Global Predictor    & \multicolumn{1}{c|}{global predictor size}                                                                                   & 2048/4096/8192                                                     \\ \hline
RAS Size            & \multicolumn{1}{c|}{return address stack size}                                                                               & 16:40:2                                                            \\ \hline
BTB Size            & \multicolumn{1}{c|}{branch target buffer size}                                                                               & 1024/2048/4096                                                     \\ \hline
ROB Size            & \multicolumn{1}{c|}{reorder buffer entries}                                                                                  & 32:256:16                                                          \\ \hline
Int RF Number       & \multicolumn{1}{c|}{\begin{tabular}[c]{@{}c@{}}number of physic\\ integer registers\end{tabular}}                            & 64:256:8                                                           \\ \hline
Fp RF Number        & \multicolumn{1}{c|}{\begin{tabular}[c]{@{}c@{}}number of physical \\ floating-point registers\end{tabular}}                  & 64:256:8                                                           \\ \hline
Inst Queue          & \multicolumn{1}{c|}{\begin{tabular}[c]{@{}c@{}}number of instruction\\ queue entries\end{tabular}}                           & 16:80:8                                                            \\ \hline
Load Queue          & \multicolumn{1}{c|}{\begin{tabular}[c]{@{}c@{}}number of\\ load queue entries\end{tabular}}                                  & 20:48:4                                                            \\ \hline
Store Queue         & \multicolumn{1}{c|}{\begin{tabular}[c]{@{}c@{}}number of\\ store queue entries\end{tabular}}                                 & 20:48:4                                                            \\ \hline
IntALU              & \multicolumn{1}{c|}{number of integer ALUs}                                                                                  & 3:8:1                                                              \\ \hline
IntMultDiv          & \multicolumn{1}{c|}{\begin{tabular}[c]{@{}c@{}}number of integer\\ multipliers and dividers\end{tabular}}                    & 1:4:1                                                              \\ \hline
FpALU               & \multicolumn{1}{c|}{\begin{tabular}[c]{@{}c@{}}number of \\ floating-point ALUs\end{tabular}}                                & 1:4:1                                                              \\ \hline
FpMultDiv           & \multicolumn{1}{c|}{\begin{tabular}[c]{@{}c@{}}number of floating-point\\ multipliers and dividers\end{tabular}}             & 1:4:1                                                              \\ \hline
Cacheline           & \multicolumn{1}{c|}{cacheline size}                                                                                          & 32/64                                                              \\ \hline
L1 ICache Size      & \multicolumn{1}{c|}{size of ICache in KB}                                                                                    & 16/32/64                                                           \\ \hline
L1 ICache Assoc.    & \multicolumn{1}{c|}{\begin{tabular}[c]{@{}c@{}}associative sets of \\ ICache\end{tabular}}                                   & 2/4                                                                \\ \hline
L1 DCache Size      & \multicolumn{1}{c|}{size of DCache in KB}                                                                                    & 16/32/64                                                           \\ \hline
L1 DCache Assoc.    & \multicolumn{1}{c|}{\begin{tabular}[c]{@{}c@{}}associative sets of \\ DCache\end{tabular}}                                   & 2/4                                                                \\ \hline
L2 Cache Size       & \multicolumn{1}{c|}{size of L2 Cache in KB}                                                                                  & 128/256                                                            \\ \hline
L2 Cache Assoc.     & \multicolumn{1}{c|}{\begin{tabular}[c]{@{}c@{}}associative sets of \\ L2 Cache\end{tabular}}                                 & 2/4                                                                \\ \hline
Total size          & \multicolumn{2}{c|}{$6.89 \times 10^{35}$}                                                                                                                                                        \\ \hline
\end{tabular}
\vspace{-5pt}
\end{table}

\textbf{Evaluation Metrics.}
A DSE framework is evaluated mainly by the accuracy of the predictor and the quality of the final Pareto optimal set.

The accuracy is mainly measured by Mean Absolute Percentage Error (MAPE), which calculates the average absolute percentage difference between real and predicted values. The formula for MAPE is:
\begin{equation}
    \text { MAPE }=\frac{1}{n} \sum_{i=1}^{n}\left(\frac{\left|y_{i}-y_{i}^{\text {real }}\right|}{y_{i}^{\text {real }}}\right) \times 100.
\end{equation}
where $n$ is the number of predicted design points, $y_i$ is the prediction objective value, i.e., IPC, power, or area in corresponding models, and $y_{i}^{real}$ is the real value; 
MAPE expresses the prediction error as a percentage of the actual values. A lower MAPE indicates better accuracy.

The PHV metric assesses the quality of the Pareto optimal set by quantifying the volume of the objective space dominated by non-dominated solutions. The PHV can be formulated as:
\begin{equation}
    \mathrm{PHV}(\Omega)=\int_\mathrm{X}\mathbb{1}[\boldsymbol{x} \prec \boldsymbol{x_{ref}}][1-\prod_{\boldsymbol{x}^* \in \Omega}\mathbb{1}[\boldsymbol{x}^* \nprec \boldsymbol{x}]]\mathrm{d}\boldsymbol{x}.
\end{equation}
Where $\mathbb{1}(·)$ is the indicator function, which outputs one if its argument is true and zero otherwise, specifically, if $x$ is the newly discovered design that is not dominated by other design points, then the PHV increases. 
See Fig.~\ref{fig:PHV_expend} for an example.



\section{Experiment Results}

\subsection{Overall Analysis}\label{sec''overall_analysis}

We evaluate the AttentionDSE framework using SPEC CPU 2017 workloads. 

\subsubsection{Comparison on Pareto Optimal Set Exploration}

\begin{figure}[!htbp] 
	
	\centering
	\includegraphics[width=0.44\textwidth]{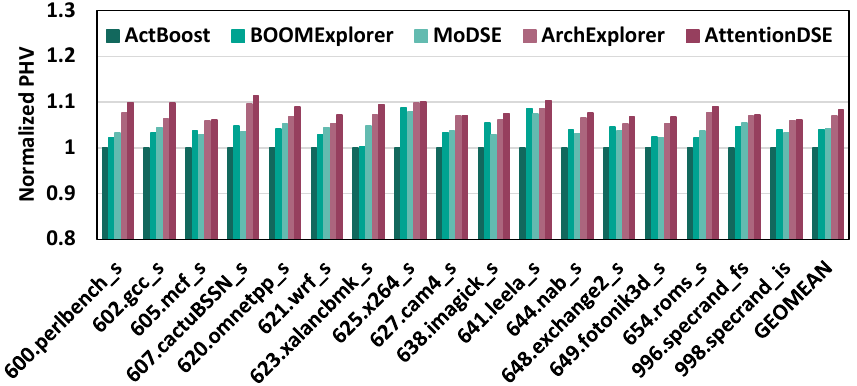}
	
	\caption{Comparison of PHV from IPC-Power optimization with SOTAs.}
	\label{fig:comparison_of_PHV}
    \vspace{-5pt}
\end{figure}

\begin{figure*}[!htbp] 
	\centering
	\includegraphics[width=0.95\textwidth]{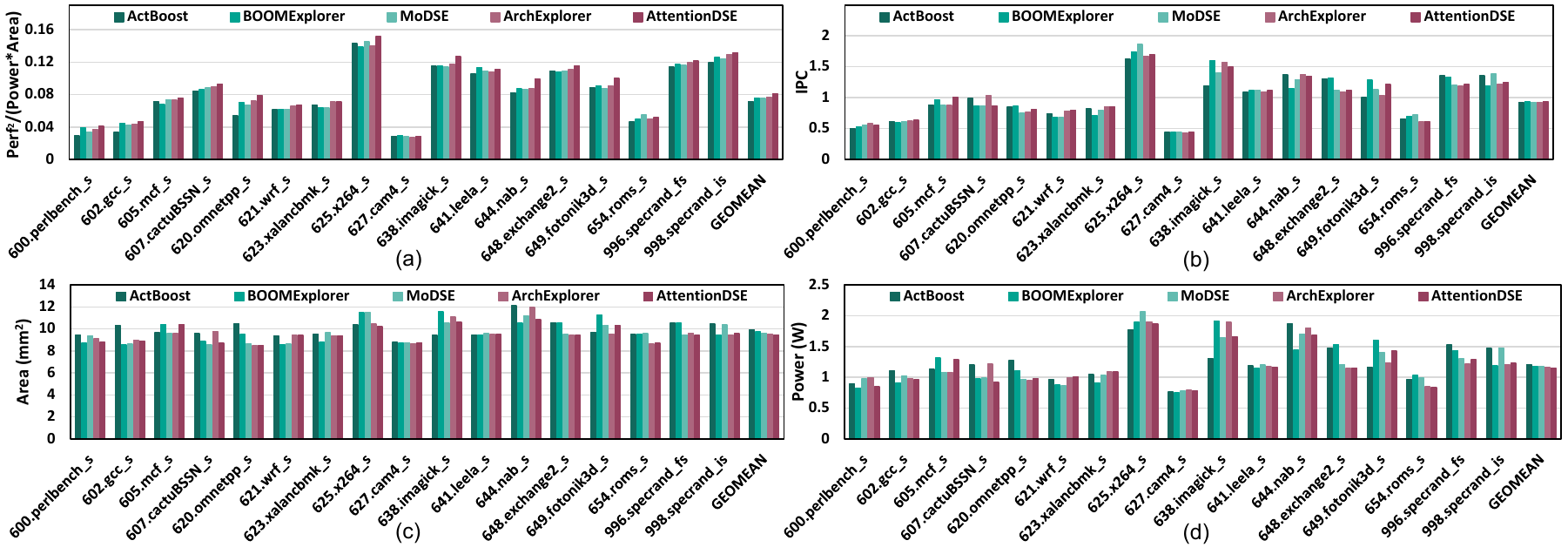}
	
	\caption{Comparisons between the Pareto designs in (a) $Perf^2/(Power \times Area)$, (b) performance, (c) area, and (d) power. }
	\label{fig:comparison_of_design_points}
    \vspace{-5pt}
\end{figure*}

Fig.~\ref{fig:comparison_of_PHV} provides a more detailed comparison of the Pareto optimal set exploration through different DSE frameworks.
Compared to SOTA DSE frameworks, AttentionDSE achieves an average improvement in PHV of 8.3\%, 4.1\%, and 3.9\% across all workloads over ActBoost, BOOMExplorer, and MoDSE, respectively. ArchExplorer employs a similar concept to AttentionDSE, analyzing program bottlenecks and addressing them accordingly. It uses DEG to identify micro-architecture bottlenecks, a process that can take several days. AttentionDSE outperforms ArchExplorer with a 1.3\% improvement in PHV. This superiority is due to two main factors: First, AttentionDSE uses an attention mechanism for design point updates and bottleneck identification, accurately pinpointing issues in the architecture. Second, ArchExplorer's bottleneck analysis is limited by complexity constraints, as it only examines the first hundred thousand lines of code. This partial analysis fails to provide a comprehensive understanding of the program, leading to reduced performance.
Note that MoDSE's exploration using the HVI approach, which searches the entire set to find the next design point, is feasible only in small design spaces. In this experiment, we use a random sampling method for design point search in MoDSE.

\begin{table}[!htbp]
\centering
\caption{Detailed information of Pareto optimal set search.}

\label{tab:detialed_information_of_DSE}
\fontsize{8pt}{8pt}\selectfont
\begin{tabular}{cclcc}
\toprule
\textbf{\begin{tabular}[c]{@{}c@{}}Methods \\ \textbackslash{}Metrics\end{tabular}} & \multicolumn{2}{c}{\textbf{PHV}} & \textbf{\begin{tabular}[c]{@{}c@{}}Up to 99\% \\ PHV Iterations\end{tabular}} & \textbf{\begin{tabular}[c]{@{}c@{}}Time\\ (Minutes)\&(Ratio)\end{tabular}} \\ 
\midrule
\textbf{BOOMExplorer} & \multicolumn{2}{c}{36.94} & 64 & 47 (9.4$\times$) \\ \midrule
\textbf{MoDSE} & \multicolumn{2}{c}{37.04} & 87 & 16  (3.2$\times$)\\ \midrule
\textbf{ArchExplorer} & \multicolumn{2}{c}{38.57} & \textbackslash{} & 828 (165.6$\times$) \\ \midrule
\textbf{AttentionDSE} & \multicolumn{2}{c}{39.41} & 15 & 5  (1$\times$)\\ 
\bottomrule  
\end{tabular}
\par Note: The experimental results are obtained from IPC-Power optimization on the 600.perlbench\_s dataset.
\vspace{-5pt}
\end{table}

Table~\ref{tab:detialed_information_of_DSE} shows the exploration results on the 600.perlbench\_s dataset across the SOTA DSE frameworks. The ``Time'' represents the total exploration time of the DSE task. All DSE frameworks, except ArchExplorer, perform a hundred iterations for the Pareto optimal set search. The ``Up to 99\% PHV Iteration'' metric indicates the convergence time of the search, where lower values are better. Experimental results demonstrate that AttentionDSE is 166 $\times$ faster than ArchExplorer, owing to ArchExplorer's time-consuming instruction analysis method. Compared to regression-based models, AttentionDSE completes the exploration roughly 10$\times$ faster than BOOMExplorer. Even with MoDSE's use of random sampling, AttentionDSE achieves exploration speeds approximately 3$\times$ faster. Additionally, AttentionDSE converges in fewer iterations, reducing iterations by 75\% compared to BOOMExplorer and 83\% compared to MoDSE, highlighting the efficiency of the ABA algorithm.

Overall, the results demonstrate that AttentionDSE provides significant improvements over existing work, validating the effectiveness of its novel methodologies in DSE.

\begin{figure*}[!htbp] 
	\centering
	\includegraphics[width=0.95\textwidth]{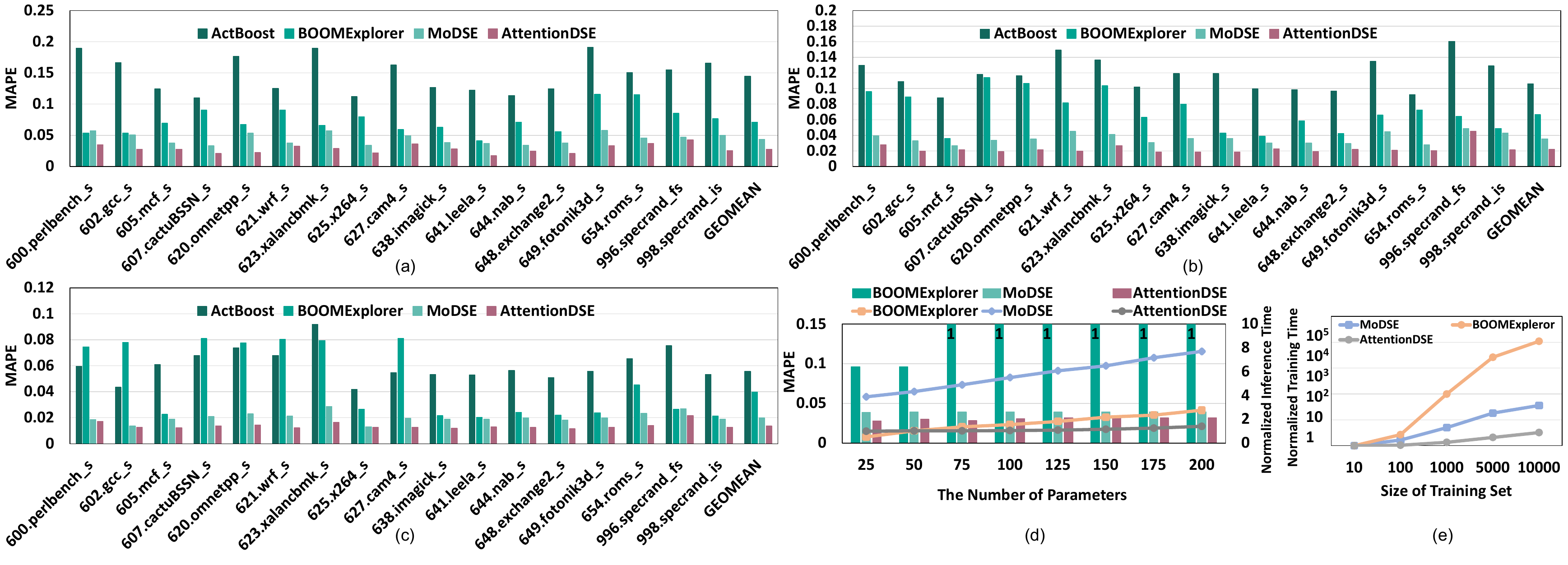}
	
	\caption{Comparison of MAPE with SOTA DSE frameworks. (a) MAPE of IPC prediction; (b) MAPE of power prediction; (c) MAPE of area prediction; (d) MAPE and normalized inference time with the increased number of parameters on the 600.perlbench\_s dataset; (e) The training time changes with the increase in training set size on the 600.perlbench\_s dataset.}
	\label{fig:MAPE}
    \vspace{-5pt}
\end{figure*}

\subsubsection{Comparison on Best Balanced Designs}

To study how high-performance designs balance PPA effectively, we select the designs with the highest performance for comparison. The metric we used for this evaluation is $Perf^2/(Power \times Area)$, which assesses how well each design balances performance with resource constraints. Fig.~\ref{fig:comparison_of_design_points} gives the results for SPEC CPU 2017, comparing frameworks such as ActBoost, BOOMExplorer, MoDSE, ArchExplorer, and AttentionDSE, with the last bar representing the geometric mean.

Across all metrics, AttentionDSE achieves, on average, the highest performance while maintaining the lowest energy consumption and area overhead, demonstrating the superiority of our methodology. Specifically, as shown in Fig.~\ref{fig:comparison_of_design_points}(a), AttentionDSE outperforms the other methods by an average of 13.8\%, 6.9\%, 7.8\%, and 5.6\%. Although AttentionDSE does not achieve the highest performance for every workload, as shown in Fig.~\ref{fig:comparison_of_design_points}(b), it consistently maintains multi-objective optimality, as shown in Fig.~\ref{fig:comparison_of_design_points}(c) and Fig.~\ref{fig:comparison_of_design_points}(d), leading to overall optimality across the design space.

\subsubsection{Comparison on Single-objective Predictor}

An accurate predictor is fundamental to an effective DSE framework. Fig.~\ref{fig:MAPE} illustrates the comparison of the prediction models with the SOTA DSE frameworks.
Fig.~\ref{fig:MAPE}(a), (b), and (c) present the MAPE of AttentionDSE compared with three DSE frameworks. 
All the predictors are trained with one thousand design points.
For example, AttentionDSE outperforms all workloads, achieving 37.1\%, 33.4\%, and 31.6\% on average for IPC, Power, and Area prediction on MAPE, respectively, compared to the SOTA DSE framework (MoDSE). 
These results highlight the superiority of AttentionDSE over statistical regression methods.
We also conduct experiments using an MLP regression model, as shown in Table~\ref{tb:detail_compare_predictor}, to demonstrate the superiority of the attention mechanism. 

\begin{table}[!htbp]
\vspace{-5pt}
\centering
\fontsize{8pt}{8pt}\selectfont
\caption{Comparison of AttentionDSE with the MLP model.}

\label{tb:detail_compare_predictor}
\begin{tabular}{ccccccc}
\toprule
\multirow{2}{*}{\textbf{\begin{tabular}[c]{@{}c@{}}Methods \\ \textbackslash{}Metrics\end{tabular}}} & \multicolumn{3}{c}{\textbf{MLP}} & \multicolumn{3}{c}{\textbf{AttentionDSE}} \\ \cmidrule{2-7} 
 & IPC & Power & Area & IPC & Power & Area \\
\midrule
\textbf{MAPE} & 0.0722 & 0.0399 & 0.0274 & 0.034 & 0.028 & 0.017 \\ \midrule
\textbf{$R^2$} & 0.8755 & 0.9689 & 0.9868 & 0.97 & 0.98 & 0.99 \\ \midrule
\textbf{MSE} & 0.0042 & 0.0009 & 0.2510 & 0.00022 & 0.0017 & 0.12 \\ 
\bottomrule
\end{tabular}
\par Note: The experimental results are obtained on the 600.perlbench\_s dataset.
\vspace{-5pt}
\end{table}

To better demonstrate support for high-dimensional DSE, we carry on additional experiments on the prediction models using the 600.perlbench\_s dataset with an increasing number of parameters and varying training set sizes, as shown in Fig.~\ref{fig:MAPE}(d) and (e). 
In Fig.~\ref{fig:MAPE}(d), the bar chart represents the comparison of MAPE, while the line chart indicates the inference time. 
BOOMExplorer, which uses GPR as the prediction model, struggles to capture effective models in high-dimensional design spaces, resulting in poor generalization performance. When the number of parameters increases to 75, BOOMExplorer fails to capture the relationship between parameters and performance metrics, resulting in a MAPE of 1.
On the other hand, although MoDSE shows an acceptable range of MAPE changes, its inference time increases sharply with the number of parameters, making it impractical for high-dimensional DSE.
Fig.~\ref{fig:MAPE}(e) illustrates the training time of each prediction model as it changes with the increase in training set size. High-dimensional design spaces need more training data to capture the relationship between micro-architecture and performance metrics. However, prevailing DSE frameworks face a significant and unacceptable increase in training time as the amount of training data grows.
For AttentionDSE, the training time remains stable due to several aspects: first, the predictor in AttentionDSE consists of the minimal necessary components for an attention-based model, ensuring lower training time through its lightweight design; second, AttentionDSE adopts the PDA mechanism, further decreasing computational complexity.

\subsection{Optimization Effect Analysis}
We further conduct an ablation experiment to demonstrate the effectiveness of the optimizations. We select 600.perlbench\_s as the representative dataset, and the results are consistent across other datasets.

\subsubsection{Efficiency of Perception-driven Attention Mechanism}

To demonstrate the benefits of the PDA mechanism, we conduct experiments to evaluate the training time and memory footprint with varying numbers of parameters, highlighting the PDA mechanism's superiority in high-dimensional DSE tasks. We set the model depth to ten and observe the impact on both training time and memory usage.

\begin{figure}[!htbp] 
    \vspace{-10pt}
	\centering
	\includegraphics[width=0.42\textwidth]{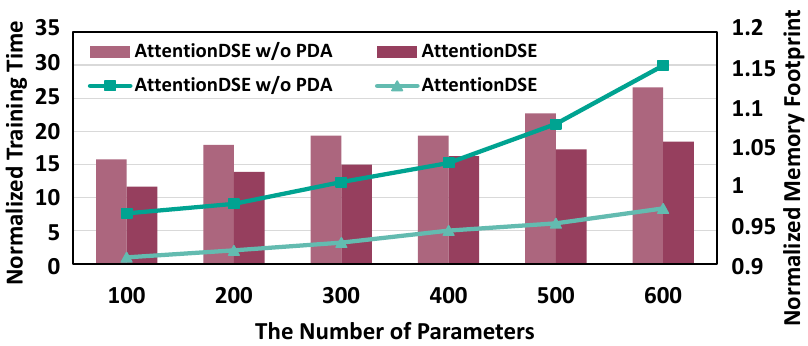}
	
	\caption{The efficiency of PDA mechanism.}
	\label{fig:PDS_efficient}
\end{figure}

Fig.~\ref{fig:PDS_efficient} shows the experiment results, the line chart illustrates the normalized training time, while the bar chart depicts the memory footprint.
The training time for models without the PDA mechanism increases exponentially with the size of the design space. For example, when the design space expands sixfold, the training time increases by 30$\times$. 
In contrast, AttentionDSE with the PDA mechanism exhibits relatively stable memory usage as the number of parameters grows. This stability extends to large design spaces because the PDA mechanism only calculates within clusters that contain the most relevant parameters.
Regarding memory footprint, in smaller design spaces, memory usage is primarily influenced by fixed data occupation during training. However, attention weights and other learnable parameters in larger design spaces gradually dominate memory usage.

\begin{figure}[!htbp] 
    \vspace{-5pt}

	\centering
	\includegraphics[width=0.42\textwidth]{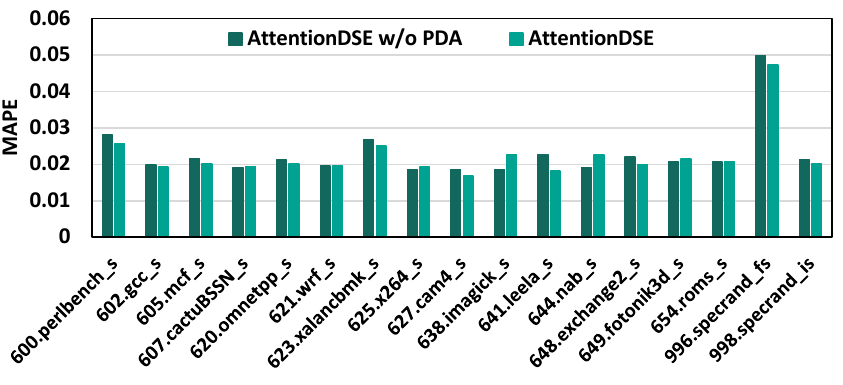}
	
	\caption{The impact of the PDA mechanism.}
	\label{fig:MAPE_difference}
\end{figure}

Fig.~\ref{fig:MAPE_difference} shows the difference in MAPE introduced by the PDA mechanism. Due to the PDA mechanism clustering the relevant parameters, captured by the sliding window attention technique, the predictive accuracy does not decrease and may even slightly increase. This improvement is primarily due to the enhanced understanding of local information within the pipeline stage provided by PDA. 
By focusing on parameter interactions within a defined window, PDA captures relationships between parameters, improving prediction accuracy.


\subsubsection{Analysis on Attention-aware Bottleneck Analysis Algorithm}

\begin{figure}[!t] 
	\centering
	\includegraphics[width=0.42\textwidth]{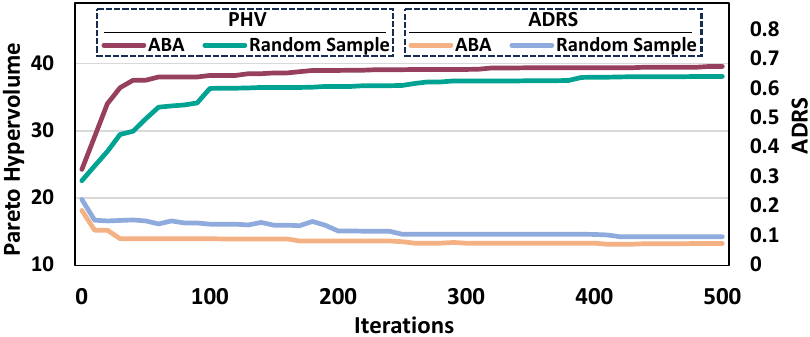}
	
	\caption{The convergence of ABA algorithm for IPC-Power optimization.}
	\label{fig:HEA_convergence}
    \vspace{-10pt}
\end{figure}

The superiority of AttentionDSE derives from both its high-accuracy prediction model and the ABA algorithm. To more effectively illustrate the impact of the ABA algorithm and avoid interference from the inaccurate prediction model, we conduct experiments on AttentionDSE by substituting the ABA algorithm with other design point search methods.
Given the large design space, exhaustive methods like MoDSE are impractical.
Since BOOMExplorer-like methods apply only to GPR models, we use random sample search as the baseline to compare with ABA. Fig.~\ref{fig:HEA_convergence} shows PHV and ADRS curves over iterations against this baseline.

Regarding PHV convergence, the ABA algorithm notably achieves a higher PHV early on, consistently surpassing the random sample search method throughout various simulation iterations. Specifically, the ABA algorithm reaches a PHV of 38.2 within one hundred iterations, while the random sample search method requires more than 5$\times$ as many iterations to achieve similar levels. This efficiency results from the ABA algorithm's effective analysis of parameter interactions and their impact on performance metrics, significantly reducing exploration time by 80\% and guiding optimal parameter updates for enhanced performance.

Concerning ADRS convergence, it serves to evaluate the quality of estimated Pareto optimal sets in multi-objective optimization by comparing them against the true Pareto set, reflecting the distribution of searched Pareto optimal design points. The ABA algorithm exhibits smoother curves, indicating a more uniform distribution of design points that cover diverse design requirements. In contrast, the random sample search method faces challenges in promptly reaching optimal intervals due to insufficient information on defects in current design points and challenges in selecting subsequent design points based on data distribution.

\section{Conclusion}



In this paper, we propose AttentionDSE, the first attention-driven framework that unifies performance prediction and architectural exploration for high-dimensional CPU design. By leveraging transformer-based attention, it achieves accurate PPA estimation and exposes parameter-wise bottlenecks to guide efficient design refinement. Through PDA and ABA, AttentionDSE delivers faster convergence and better Pareto quality than prior methods. This work demonstrates the potential of attention mechanisms as a scalable and interpretable solution for automated micro-architecture design.



\bibliographystyle{IEEEtranS}
\bibliography{refs}

\end{document}